%% file: main-1720-govindarajan.tex
\documentclass[11pt,a4paper]{article}
\usepackage{times,latexsym}
\usepackage{url}
\usepackage[T1]{fontenc}

\usepackage[acceptedWithA]{tacl2018v2}

\usepackage{xspace,mfirstuc,tabulary}

\newcommand{\ex}[1]{{\sf #1}}

\newif\iftaclinstructions
\taclinstructionsfalse 
\iftaclinstructions
\renewcommand{\confidential}{}
\renewcommand{\anonsubtext}{(No author info supplied here, for consistency with
TACL-submission anonymization requirements)}
\newcommand{\instr}
\fi

\iftaclpubformat 

\else

\fi

\usepackage[utf8]{inputenc}
\usepackage{enumitem}
\setlist{nosep}
\usepackage{styledefs}
\usepackage{booktabs}
\usepackage{linguex}
\usepackage{colortbl}
\usepackage{multirow}
\usepackage{rotating}
\usepackage{diagbox}
\usepackage{float}
\usepackage{stfloats}

\title{Decomposing Generalization\\{\normalsize Models of Generic, Habitual, and Episodic Statements}}

\author{Venkata Govindarajan\\University of Rochester \And Benjamin Van Durme\\Johns Hopkins University \And Aaron Steven White\\University of Rochester}

\date{}

\begin{document}
\maketitle

\begin{abstract}
\input{sections/abstract}    
\end{abstract}

\setlength{\Exlabelsep}{0em}
\setlength{\Extopsep}{.2\baselineskip}
\setlength{\SubExleftmargin}{1.3em}

\section{Introduction}
\label{sec:introduction}
\input{sections/01-introduction}

\section{Background}
\label{sec:background}
\input{sections/02-background}

\section{Annotation Framework}
\label{sec:annotationframework}
\input{sections/03-annotationframework}

\section{Framework Validation}
\label{sec:frameworkvalidation}
\input{sections/04-frameworkvalidation}

\section{Comparison to Standard Ontology}
\label{sec:comparison}
\input{sections/05-comparison}

\section{Bulk Annotation}
\label{sec:datacollection}
\input{sections/06-bulkannotation}

\section{Exploratory Analysis}
\label{sec:preliminaryanalysis}
\input{sections/07-exploratoryanalysis}

\section{Models}
\label{sec:models}
\input{sections/08-models}

\section{Results}
\label{sec:results}
\input{sections/09-results}

\section{Analysis}
\label{sec:analysis}
\input{sections/10-analysis}

\section{Conclusion}
\label{sec:conclusion}
\input{sections/11-conclusion}

\section*{Acknowledgments}

\input{sections/acknowledgments}

\bibliography{references}
\bibliographystyle{acl_natbib}

\end{document}

%% file: sections/abstract.tex
We present a novel semantic framework for modeling linguistic expressions of generalization---\textit{generic}, \textit{habitual}, and \textit{episodic statements}---as combinations of simple, real-valued referential properties of predicates and their arguments.\ We use this framework to construct a dataset covering the entirety of the Universal Dependencies English Web Treebank.\ We use this dataset to probe the efficacy of type-level and token-level information---including hand-engineered features and static (GloVe) and contextual (ELMo) word embeddings---for predicting expressions of generalization.

%% file: sections/01-introduction.tex
Natural language allows us to convey not only information about particular individuals and events, as in \ref{ex:episodic}, but also generalizations about those individuals and events, as in \ref{ex:habitual}.

\ex. \label{ex:episodic}
\a. Mary ate oatmeal for breakfast today. \label{ex:episodic-a}
\b. The students completed their assignments. \label{ex:episodic-b}

\vspace{-6mm}

\ex.\label{ex:habitual}
\a. Mary eats oatmeal for breakfast.\label{ex:habitual-a}
\b. The students always complete their assignments on time. \label{ex:habitual-b}

This capacity for expressing generalization is extremely flexible---allowing for generalizations about the kinds of events that particular individuals are habitually involved in, as in \ref{ex:habitual}, as well as characterizations about kinds of things, as in \ref{ex:generic}.

\ex.\label{ex:generic}
\a. Bishops move diagonally. \label{ex:generic-a}
\b. Soap is used to remove dirt. \label{ex:generic-b}

Such distinctions between \textit{episodic statements} \ref{ex:episodic}, on the one hand, and \textit{habitual} \ref{ex:habitual} and \textit{generic (or characterizing) statements} \ref{ex:generic}, on the other, have a long history in both the linguistics and artificial intelligence literatures \citep[see][]{Carlson:2011, Maienborn:vonHeusinger:Portner:2011b,sep-generics}. Nevertheless, few modern semantic parsers make a systematic distinction (cf. \citealt{abzianidze2017towards}).

This is problematic, because the ability to accurately capture different modes of generalization is likely key to building systems with robust common sense reasoning \citep{zhang_ordinal_2017,bauer2018commonsense}: such systems need some source for general knowledge about the world \citep{mccarthy1960programs,mccarthy_circumscriptionform_1980,mccarthy_applications_1986,minsky_framework_1974,schank_scripts_1975,Hobbs:1987:CML:48160.48164,reiter_nonmonotonic_1987} and natural language text seems like a prime candidate. It is also surprising, since there is no dearth of data relevant to linguistic expressions of generalization \citep{doddington_automatic_2004,cybulska2014using, friedrich_annotating_2015}.

One obstacle to further progress on generalization is that current frameworks tend to take standard descriptive categories as sharp classes---e.g.\ \textsc{episodic}, \textsc{generic}, \textsc{habitual} for statements and \textsc{kind}, \textsc{individual} for noun phrases.  This may seem reasonable for sentences like \ref{ex:episodic-a}, where \textit{Mary} clearly refers to a particular individual, or  \ref{ex:generic-a}, where \textit{Bishops} clearly refers to a kind; but natural text is less forgiving \cite{grimm2014individuating,grimm_crime_2016,grimm_grammatical_2018}. Consider the underlined arguments in \ref{ex:data-intro}: do they refer to kinds or individuals?

\ex. \label{ex:data-intro}
\a. I will manage \underline{client expectations}.\label{ex:data-intro-a}
\b. \underline{The atmosphere} may not be for everyone.\label{ex:data-intro-b}
\c. Thanks again for \underline{great customer service}!\label{ex:data-intro-c}

To remedy this, we propose a novel framework for capturing linguistic expressions of generalization. Taking inspiration from \textit{decompositional semantics} \citep{reisinger2015semantic, white_universal_2016}, we suggest that linguistic expressions of generalization should be captured in a continuous multi-label system, rather than a multi-class system. We do this by decomposing categories such as \textsc{episodic}, \textsc{habitual}, and \textsc{generic} into simple referential properties of predicates and their arguments. Using this framework (\S\ref{sec:annotationframework}), we develop an annotation protocol, which we validate (\S\ref{sec:frameworkvalidation}) and compare against previous frameworks (\S\ref{sec:comparison}). We then deploy this framework (\S\ref{sec:datacollection}) to construct a new large-scale dataset of annotations covering the entire Universal Dependencies \cite{de_marneffe_universal_2014, nivre_universal_2015} English Web Treebank \citep[UD-EWT;][]{bies2012english, silveira_gold_2014}---yielding the Universal Decompositional Semantics-Genericity (UDS-G) dataset.\footnote{Data, code, protocol implementation, and task instructions provided to annotators are available at \href{http://decomp.io}{decomp.io}.} Through exploratory analysis of this dataset, we demonstrate that this multi-label framework is well-motivated (\S\ref{sec:preliminaryanalysis}). We then present models for predicting expressions of linguistic generalization that combine hand-engineered type and token-level features with static and contextual learned representations (\S\ref{sec:models}). We find that (i) referential properties of arguments are easier to predict than those of predicates; and that (ii) contextual learned representations contain most of the relevant information for both arguments and predicates (\S\ref{sec:results}).

%% file: sections/02-background.tex
Most existing annotation frameworks aim to capture expressions of linguistic generalization using multi-class annotation schemes. We argue that this reliance on multi-class annotation schemes is problematic on the basis of descriptive and theoretical work in the linguistics literature.

One of the earliest frameworks explicitly aimed at capturing expressions of linguistic generalization was developed under the \textbf{ACE-2} program \cite[][and see \citealt{reiter_identifying_2010}]{ace2,doddington_automatic_2004}. This framework associates entity mentions with discrete labels for whether they refer to a specific member of the set in question (\textsc{specific}) or any member of the set in question (\textsc{generic}). The \textbf{ACE-2005} Multilingual Training Corpus \cite{ace2005} extends these annotation guidelines, providing two additional classes: (i) negatively quantified entries (\textsc{neg}) for referring to empty sets and (ii) underspecified entries (\textsc{usp}), where the referent is ambiguous between \textsc{generic} and \textsc{specific}.

The existence of the \textsc{usp} label already portends an issue with multi-class annotation schemes, which have no way of capturing the well-known phenomena of \textit{taxonomic reference} \citep[see][and references therein]{carlson_generic_1995}, \textit{abstract/event reference} \citep{grimm2014individuating, grimm_crime_2016, grimm_grammatical_2018}, and \textit{weak definites} \citep{CarlsonEtAl06}. For example, \textit{wines} in \ref{ex:taxonomic} refers to particular kinds of wine; \textit{service} in \ref{ex:abstract} refers to an abstract entity/event that could be construed as both particular-referring, in that it is the service at a specific restaurant, and kind-referring, in that it encompasses all service events at that restaurant; and \textit{bus} in \ref{ex:weakdefnite} refers to potentially multiple distinct buses that are grouped into a kind by the fact that they drive a particular line.

\ex. That vintner makes \underline{three different wines}. \label{ex:taxonomic}

\ex. \underline{The service at that restaurant} is excellent. \label{ex:abstract}

\ex. That bureaucrat takes \underline{the 90 bus} to work. \label{ex:weakdefnite}

This deficit is remedied to some extent in the \textbf{ARRAU} \citep[][and see \citealt{mathew_supervised_2009,louis_automatic_2011}]{poesio2008anaphoric} and \textbf{ECB+} \cite{cybulska2014using,ecb2014} corpora. The ARRAU corpus is mainly intended to capture anaphora resolution, but following the GNOME guidelines \cite{poesio2004discourse}, it also annotates entity mentions for a \textsc{generic} attribute, sensitive to whether the mention is in the scope of a relevant semantic operator---e.g.\ a conditional or quantifier---and whether the nominal refers to a type of object whose genericity is left underspecified, such as a substance. The ECB+ corpus is an extension of the EventCorefBank \citep[ECB;][]{bejan2010unsupervised,lee_joint_2012}, which annotates Google News texts for event coreference in accordance with the TimeML specification \cite{pustejovsky2003timebank}, and is an improvement in the sense that, in addition to entity mentions, event mentions may be labeled with a \textsc{generic} class.

The ECB+ approach is useful, since episodic, habitual, and generic statements can straightforwardly be described using combinations of event and entity mention labels. For example, in ECB+, episodic statements involve only non-generic entity and event mentions; habitual statements involve a generic event mention and at least one non-generic entity mention; and generic statements involve generic event mentions and at least one generic entity mention. This demonstrates the strength of decomposing statements into properties of the events and entities they describe; but there remain difficult issues arising from the fact that the decomposition does not go far enough. One is that, like ACE-2/2005 and ARRAU, ECB+ does not make it possible to capture taxonomic and abstract reference or weak definites; another is that, because ECB+ treats generics as mutually exclusive from other event classes, it is not possible to capture that events and states in those classes can themselves be particular or generic. This is well-known for different classes of events, such as those determined by a predicate's \textit{lexical aspect} \citep{vendler_verbs_1957}; but it is likely also important for distinguishing more particular \textit{stage-level properties}---e.g.\ availability \ref{ex:stagelevel}---from more generic \textit{individual-level properties}---e.g.\ strength \ref{ex:individuallevel} \citep{carlson1977reference}.

\ex. Those firemen \underline{are available}. \label{ex:stagelevel}

\ex. Those firemen \underline{are strong}. \label{ex:individuallevel}

This situation is improved upon in the Richer Event Descriptions \citep[\textbf{RED};][]{ogorman_richer_2016} and Situation Entities \citep[\textbf{SitEnt}; ][]{friedrich2014automatic,friedrich_situation_2014,friedrich_annotating_2015,friedrich2015discourse,friedrich2015automatic,friedrich_situation_2016} frameworks, which annotate both NPs and entire clauses for genericity. In particular, SitEnt, which is used to annotate MASC \cite{ide2010manually} and Wikipedia, has the nice property that it recognizes the existence of abstract entities and lexical aspectual class of clauses' main verbs, along with habituality and genericity. This is useful because, in addition to decomposing statements using the genericity of the main referent and event, this framework recognizes that lexical aspect is an independent phenomenon. In practice, however, the annotations produced by this framework are mapped into a multi-class scheme containing only the high-level \textsc{generic}-\textsc{habitual}-\textsc{episodic} distinction---alongside a conceptually independent distinction among illocutionary acts.

A potential argument in favor of mapping into a multi-class scheme is that, if it is sufficiently elaborated, the relevant decomposition may be recoverable. But regardless of such an elaboration, uncertainty about which class any particular entity or event falls into cannot be ignored. Some examples may just not have categorically correct answers; and even if they do, annotator uncertainty and bias may obscure them. To account for this, we develop a novel annotation framework that both (i) explicitly captures annotator confidence about the different referential properties discussed above and (ii) attempts to correct for annotator bias using standard psycholinguistic methods.

%% file: sections/03-annotationframework.tex
We divide our framework into two protocols---the \textit{argument} and \textit{predicate protocols}---that probe properties of individuals and situations---i.e.\ events or states---referred to in a clause. Drawing inspiration from prior work in \textit{decompositional semantics} \citep{white_universal_2016}, a crucial aspect of our framework is that (i) multiple properties can be simultaneously true for a particular individual or situation (event or state); and (ii) we explicitly collect confidence ratings for each property. This makes our framework highly extensible, since further properties can be added without breaking a strict multi-class ontology.

Drawing inspiration from the prior literature on generalization discussed in \S\ref{sec:introduction} and \S\ref{sec:background}, we focus on properties that lie along three main axes: whether a predicate or its arguments refer to (i) instantiated or spatiotemporally delimited---i.e.\ \textit{particular}---situations or individuals; (ii) classes of situations---i.e.\ \textit{hypothetical} situations---or \textit{kinds} of individuals; and/or (iii) intangible---i.e.\ \textit{abstract} concepts or \textit{stative} situations.

This choice of axes is aimed at allowing our framework to capture not only the standard \textsc{episodic}-\textsc{habitual}-\textsc{generic} distinction, but also phenomena that do not fit neatly into this distinction, such as taxonomic reference, abstract reference, and weak definites. The idea here is similar to prior decompositional semantics work on \textit{semantic protoroles} \citep{reisinger2015semantic,white_universal_2016,white_semantic_2017}, which associates categories like \textsc{agent} or \textsc{patient} with sets of more basic properties, such as volitionality, causation, change-of-state, etc., and is similarly inspired by classic theoretical work \citep{dowty_thematic_1991}. 
\begin{figure}[t]
\centering
\includegraphics[width=\columnwidth]{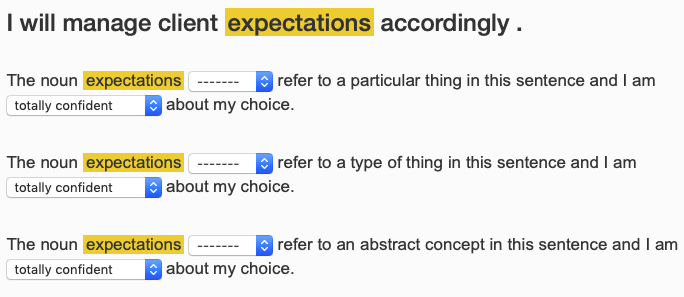}
\includegraphics[width=\columnwidth]{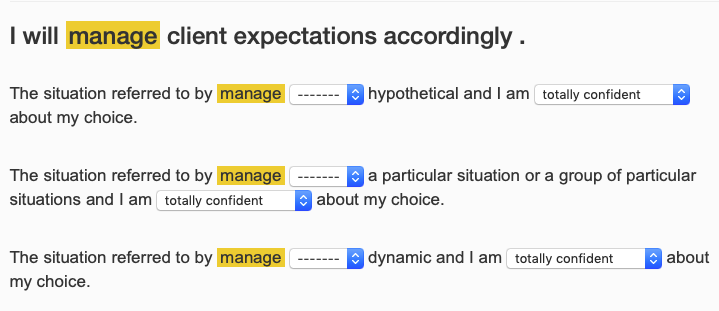}
\vspace{-7mm}
\caption{Examples of argument protocol (top) and predicate protocol (bottom).}
\label{fig:protocol}
\end{figure}

In our framework, prototypical episodics, habituals, and generics correspond to sets of properties that the referents of a clause's head predicate and arguments have---i.e.\ clausal categories are built up from properties of the predicates that head them along with those predicates' arguments. For instance, prototypical episodic statements, like those in \ref{ex:episodic}, have arguments that only refer to particular, non-kind, non-abstract individuals and a predicate that refers to a particular event or (possibly) state; prototypical habitual statements, like those in \ref{ex:habitual} have arguments that refer to at least one particular, non-kind, non-abstract individual and a predicate that refers to a non-particular, dynamic event; and prototypical generics, like those in \ref{ex:generic}, have arguments that only refer to kinds of individuals and a predicate that refers to non-particular situations.

It is important to note that these are all \textit{prototypical} properties of episodic, habitual, or generic statements, in the same way that volitionality is a prototypical property of agents and change-of-state is a prototypical property of patients. That is, our framework explicitly allows for bleed between categories because it only commits to the referential properties, not the categories themselves. It is this ambivalence toward sharp categories that also allows our framework to capture phenomena that fall outside the bounds of the standard three-way distinction. For instance, taxonomic reference, as in \ref{ex:taxonomic}, and weak definites, as in \ref{ex:weakdefnite}, prototypically involve an argument being both particular- and kind-referring; stage-level properties, as in \ref{ex:stagelevel}, prototypically involve particular, non-dynamic situations, while individual-level properties, as in \ref{ex:individuallevel}, prototypically involve non-particular, non-dynamic situations.

Figure \ref{fig:protocol} shows examples of the argument protocol (top) and predicate protocol (bottom), whose implementation is based on the event factuality annotation protocol described by \citet{white_universal_2016} and \citet{rudinger_neural_2018}. Annotators are presented with a sentence with one or many words highlighted, followed by statements pertaining to the highlighted words in the context of the sentence. They are then asked to fill in the statement with a binary response saying whether it \textit{does} or \textit{does not} hold and to give their confidence on a 5 point scale---\textit{not at all confident} (1), \textit{not very confident} (2), \textit{somewhat confident} (3), \textit{very confident} (4), and \textit{totally confident} (5).

%% file: sections/04-frameworkvalidation.tex
To demonstrate the efficacy of our framework for use in bulk annotation (reported in \S\ref{sec:datacollection}), we conduct a validation study on both our predicate and argument protocols. The aim of these studies is to establish that annotators display reasonable agreement when annotating for the properties in each protocol, relative to their reported confidence. We expect that, the more confident both annotators are in their annotation, the more likely it should be that annotators agree on those annotations.

To ensure that the findings from our validation studies generalize to the bulk annotation setting, we simulate the bulk setting as closely as possible: (i) randomly sampling arguments and predicates for annotation from the same corpus we conduct the bulk annotation on (UD-EWT); and (ii) allowing annotators to do as many or as few annotations as they would like. This design makes standard measures of interannotator agreement somewhat difficult to accurately compute, since different pairs of annotators may annotate only a small number of overlapping items (arguments/predicates), so we turn to standard statistical methods from psycholinguistics to assist in estimation of interannotator agreement.

\vspace{-1mm}

\paragraph{Predicate and argument extraction} We extracted predicates and their arguments from the gold UD parses from UD-EWT using PredPatt \cite{white_universal_2016,zhang_evaluation_2017}. From the UD-EWT training set, we then randomly sampled 100 arguments from those headed by a \texttt{DET}, \texttt{NUM}, \texttt{NOUN}, \texttt{PROPN}, or \texttt{PRON} and 100 predicates from those headed by a \texttt{ADJ}, \texttt{NOUN}, \texttt{NUM}, \texttt{DET}, \texttt{PROPN}, \texttt{PRON}, \texttt{VERB}, or \texttt{AUX}.

\vspace{-1mm}

\paragraph{Annotators} 44 annotators were recruited from Amazon Mechanical Turk to annotate arguments; and 50 annotators were recruited to annotate predicates. In both cases, arguments and predicates were presented in batches of 10, with each predicate and argument annotated by 10 annotators.

\vspace{-1mm}

\paragraph{Confidence normalization} Because different annotators use the confidence scale in different ways---e.g.\ some annotators use all five options while others only ever respond with \textit{totally confident} (5)----we normalize the confidence ratings for each property using a standard ordinal scale normalization technique known as ridit scoring \citep{agresti_categorical_2003}. In ridit scoring, ordinal labels are mapped to (0, 1) using the empirical cumulative distribution function of the ratings given by each annotator. Specifically, for the responses $\yb^{(a)}$ given by annotator $a$, $\text{ridit}_{\yb^{(a)}}\left(y^{(a)}_i\right) = \text{ECDF}_{\yb^{(a)}}\left(y^{(a)}_i-1\right) + 0.5 \times \text{ECDF}_{\yb^{(a)}}\left(y^{(a)}_i\right)$. 

Ridit scoring has the effect of reweighting the importance of a scale label based on the frequency with which it is used. For example, insofar as an annotator rarely uses extreme values, such as \textit{not at all confident} or \textit{totally confident}, the annotator is likely signaling very low or very high confidence, respectively, when they are used; and insofar as an annotator often uses extreme values, the annotator is likely not signaling particularly low or particularly high confidence.

\begin{table}[t]
    \centering
    \footnotesize
    \input{tables/bias.tex}
    \caption{Bias (log-odds) for answering \textit{true}}
    \label{tab:biases}
\end{table}

\paragraph{Interannotator Agreement (IAA)} Common IAA statistics, such as Cohen's or Fleiss' $\kappa$, rely on the ability to compute both an expected agreement $p_e$ and an observed agreement $p_o$, with $\kappa \equiv \frac{p_o - p_e}{1 - p_ e}$. Such a computation is relatively straightforward when a small number of annotators annotate many items, but when many annotators each annotate a small number of items pairwise, $p_e$ and $p_o$ can be difficult to estimate accurately, especially for annotators that only annotate a few items total. Further, there is no standard way to incorporate confidence ratings as we have collected them into these IAA measures. 

To overcome these obstacles, we use a combination of mixed and random effects models \citep[][]{gelman_data_2014}, which are extremely common in the analysis of psycholinguistic data \citep{baayen_analyzing_2008}, to estimate $p_e$ and $p_o$ for each property. The basic idea behind using these models is to allow our estimates of $p_e$ and $p_o$ to be sensitive to the number of items annotators annotated as well as how annotators' confidence relates to agreement.

To estimate $p_e$ for each property, we fit a random effects logistic regression to the binary responses for that property, with random intercepts for both annotator and item. The fixed intercept estimate $\hat{\beta}_0$ for this model is an estimate of the log-odds that the average annotator would answer \textit{true} on that property for the average item; and the random intercepts give the distribution of actual annotator ($\hat{\sigma}_\text{ann}$) or item ($\hat{\sigma}_\text{item}$) biases. Table \ref{tab:biases} gives the estimates for each property. We note a substantial amount of variability in the bias different annotators have for answering true on many of these properties. This variability is evidenced by the fact that $\hat{\sigma}_\text{ann}$ and $\hat{\sigma}_\text{item}$ are similar across properties, and it suggests the need to adjust for annotator biases when analyzing these data, which we do both here and for our bulk annotation.

To compute $p_e$ from these estimates, we use a parametric bootstrap. On each replicate, we sample annotator biases $b_1, b_2$ independently from $\mathcal{N}(\hat{\beta}_0, \hat{\sigma}_\text{ann})$, then compute the expected probability of random agreement in the standard way: $\pi_1\pi_2 + (1-\pi_1)(1-\pi_2)$, where $\pi_i = \text{logit}_{-1}(b_i)$. We compute the mean across 9,999 such replicates to obtain $p_e$, shown in Table \ref{tab:iaa}.

\begin{table}[t]
    \centering
    \footnotesize
    \input{tables/iaa.tex}
    \caption{Interannotator agreement scores}
    \label{tab:iaa}
\end{table}

To estimate $p_o$ for each property in a way that takes annotator confidence into account, we first compute, for each pair of annotators, each item they both annotated, and each property they annotated that item on, whether or not they agree in their annotation. We then fit separate mixed effects logistic regressions for each property to this agreement variable, with a fixed intercept $\beta_0$ and slope $\beta_\text{conf}$ for the product of the annotators' confidence for that item and random intercepts for both annotator and item.\footnote{We use the product of annotator confidences because it is large when both annotators have high confidence and small when either annotator has low confidence and always remains between 0 (lowest confidence) and 1 (highest confidence).}

We find, for all properties, that there is a reliable increase---i.e.\ a positive $\hat{\beta}_\text{conf}$---in agreement as annotators' confidence ratings go up ($p$s $< 0.001$). This corroborates our prediction that annotators should have higher agreement for things they are confident about. It also suggests the need to incorporate confidence ratings into the annotations our models are trained on, which we do in our normalization of the bulk annotation responses.

From the fixed effects, we can obtain an estimate of the probability of agreement for the average pair of annotators at each confidence level between 0 and 1. We compute two versions of $\kappa$ based on such estimates: $\kappa_\text{low}$, which corresponds to 0 confidence for at least one annotator in a pair, and $\kappa_\text{high}$, which corresponds to perfect confidence for both. Table \ref{tab:iaa} shows these $\kappa$ estimates. 

As implied by reliably positive $\hat{\beta}_\text{conf}$s, we see that $\kappa_\text{high}$ is greater than $\kappa_\text{low}$ for all properties. Further, with the exception of \textsc{dynamic}, $\kappa_\text{high}$ is generally comparable to the $\kappa$ estimates reported in annotations by trained annotators using a multi-class framework. For instance, compare the metrics in Table \ref{tab:iaa} to $\kappa_\text{ann}$ in Table \ref{tab:comparison} (see \S\ref{sec:comparison} for details), which gives the Fleiss' $\kappa$ metric for clause types in the SitEnt dataset \citep{friedrich_situation_2016}.

%% file: tables/bias.tex
\begin{tabular}{clrrr}
	\toprule
    & \textbf{Property} & $\hat{\beta}_0$ & $\hat{\sigma}_\text{ann}$ & $\hat{\sigma}_\text{item}$\\
    \midrule
    {\multirow{3}{*}{\rotatebox{90}{\tiny Argument}}} & Is.Particular & 0.49 & 1.15 & 1.76\\
    & Is.Kind & -0.31 & 1.23 & 1.34\\
    & Is. Abstract & -1.29 & 1.27 & 1.70\\
    \midrule
    {\multirow{3}{*}{\rotatebox{90}{\tiny Predicate}}} & Is.Particular & 0.98 & 0.91 & 0.72\\
    & Is.Dynamic & 0.24 & 0.82 & 0.59\\
    & Is.Hypothetical & -0.78 & 1.24 & 0.90\\
    \bottomrule
\end{tabular}

%% file: tables/iaa.tex
\begin{tabular}{clrrr}
	\toprule
    & \textbf{Property} & $p_e$ &  $\kappa_\text{low}$ & $\kappa_\text{high}$\\
    \midrule
    {\multirow{3}{*}{\rotatebox{90}{\tiny Argument}}} & Is.Particular & 0.52 & 0.21 & 0.77 \\
    & Is.Kind & 0.51 & 0.12 & 0.51\\
    & Is. Abstract & 0.61 & 0.17 & 0.80\\
    \midrule
    {\multirow{3}{*}{\rotatebox{90}{\tiny Predicate}}} & Is.Particular & 0.58 & -0.11 & 0.54 \\
    & Is.Dynamic & 0.51 & -0.02 & 0.22\\
    & Is.Hypothetical & 0.54 & -0.04 & 0.60 \\
    \bottomrule
\end{tabular}



%% file: sections/05-comparison.tex
To demonstrate that our framework subsumes standard distinctions---e.g.\ \textsc{episodic} v. \textsc{habitual} v. \textsc{generic}---we conduct a study comparing annotations assigned under our multi-label framework to those assigned under a framework that recognizes such multi-class distinctions. We choose the the SitEnt framework for this comparison, since it assumes a categorical distinction between \textsc{generic}, \textsc{habitual} (their \textsc{generalizing}), \textsc{episodic} (their \textsc{eventive}), and \textsc{stative} clauses \citep{friedrich2014automatic,friedrich_situation_2014,friedrich_annotating_2015,friedrich2015discourse,friedrich2015automatic,friedrich_situation_2016}.\footnote{SitEnt additionally assumes three other classes, contrasting with the four above: \textsc{imperative}, \textsc{question}, and \textsc{report}. We ignore clauses labeled with these categories.} SitEnt is also a useful comparison because it was constructed by highly trained annotators who had access to the entire document containing the clause being annotated, thus allowing us to assess both how much it matters that we use only very lightly trained annotators and do not provide document context. 

\vspace{-1mm}

\paragraph{Predicate and argument extraction} For each of \textsc{generic}, \textsc{habitual}, \textsc{stative}, and \textsc{eventive}, we randomly sample 100 clauses from SitEnt such that (i) that clause's gold annotation has that category; and (ii) all SitEnt annotators agreed on that annotation. We annotate the \texttt{mainReferent} of these clauses (as defined by SitEnt) in our argument protocol and the \texttt{mainVerb} in our predicate protocol, providing annotators only the sentence containing the clause.

\paragraph{Annotators} 42 annotators were recruited from Amazon Mechanical Turk to annotate arguments, and 45 annotators were recruited to annotate predicates---both in batches of 10, with each predicate and argument annotated by 5 annotators.

\begin{table}[t]
    \centering
    \footnotesize
    \input{tables/comparison.tex}
    \caption{Predictability of standard ontology using our property set in a kernelized support vector classifier.}
    \label{tab:comparison}
\end{table}

\vspace{-1mm}

\paragraph{Annotation normalization} As noted in \S\ref{sec:frameworkvalidation}, different annotators use the confidence scale differently and have different biases for responding \textit{true} or \textit{false} on different properties (see Table \ref{tab:biases}). To adjust for these biases, we construct a normalized score for each predicate and argument using mixed effects logistic regressions. These mixed effects models all had (i) a hinge loss with margin set to the normalized confidence rating; (ii) fixed effects for property---\textsc{particular}, \textsc{kind}, and \textsc{abstract} for arguments; \textsc{particular}, \textsc{hypothetical}, and \textsc{dynamic} for predicates---token, and their interaction; and (iii) by-annotator random intercepts and random slopes for property with diagonal covariance matrices. The rationale behind (i) is that \textit{true} should be associated with positive values; \textit{false} should be associated with negative values; and the confidence rating should control how far from zero the normalized rating is, adjusting for the biases of annotators that responded to a particular item. The resulting response scale is analogous to current approaches to event factuality annotation \citep{lee_event_2015,stanovsky_integrating_2017,rudinger_neural_2018}.

\begin{figure}[t]
    \centering
    \includegraphics[width=0.92\columnwidth]{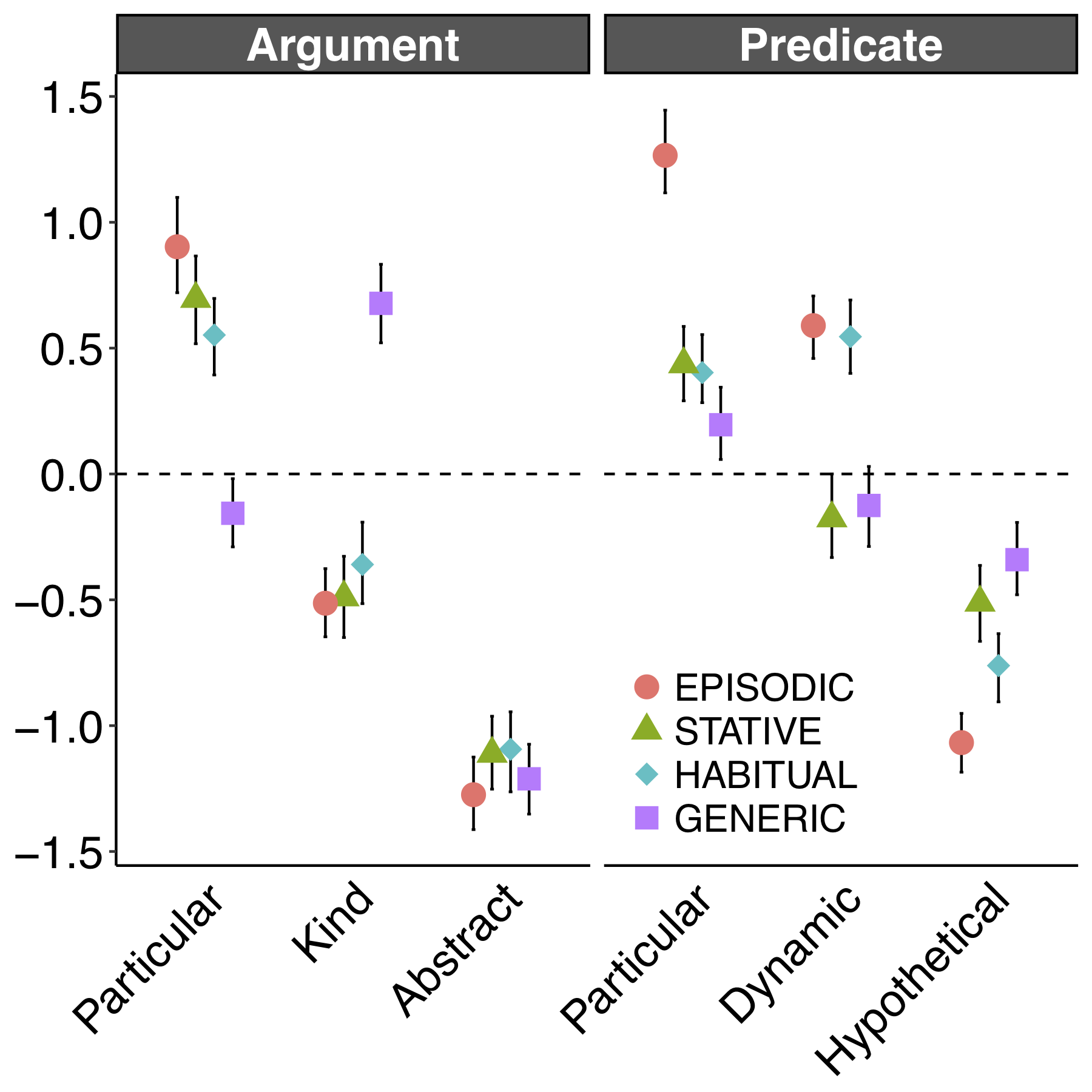}
    \vspace{-4mm}
    \caption{Mean property value for each clause type.}
    \label{fig:comparison}
\end{figure}

We obtain a normalized score from these models by setting the Best Linear Unbiased Predictors for the by-annotator random effects to zero and using the Best Linear Unbiased Estimators for the fixed effects to obtain a real-valued label for each token on each property. This procedure amounts to estimating a label for each property and each token based on the `average annotator.' 

\vspace{-1mm}

\paragraph{Quantitative comparison} To compare our annotations to the gold situation entity types from SitEnt, we train a support vector classifier (SVC) with a radial basis function (RBF) kernel to predict the situation entity type of each clause on the basis of the normalized argument property annotations for that clause's \texttt{mainReferent} and the normalized predicate property annotations for that clause's \texttt{mainVerb}. The hyperparameters for this SVC were selected using exhaustive grid search over the regularization parameter $\lambda \in$ $\left\{1, 10, 100, 1000\right\}$ and bandwidth $\sigma \in$ $\left\{10^{-2}, 10^{-3}, 10^{-4}, 10^{-5}\right\}$ in a 5-fold cross-validation (CV). This 5-fold CV was nested within a 10-fold CV, from which we calculate metrics.

Table \ref{tab:comparison} reports the precision, recall, and F-score computed using the held-out set in each fold of the 10-fold CV. For purposes of comparison, it also gives the Fleiss' $\kappa$ reported by \citet{friedrich_situation_2016} for each property ($\kappa_\text{ann}$) as well as Cohen's $\kappa$ between our model predictions on the held-out folds and the gold SitEnt annotations ($\kappa_\text{mod}$). One way to think about $\kappa_\text{mod}$ is that it tells us what agreement we would expect if we used our model as an annotator instead of highly trained humans.

We see that our model's agreement ($\kappa_\text{mod}$) tracks interannotator agreement ($\kappa_\text{ann}$) surprisingly well. Indeed, in some cases, such as for \textsc{generic}, our model's agreement is within a few points of interannotator agreement. This pattern is surprising, since our model is based on annotations by very lightly trained annotators who have access to very limited context compared to the annotators of SitEnt, who get the entire document a clause is found in. Indeed, our model has access to even less context than it could otherwise have on the basis of our framework, since we only annotate one of the potentially many arguments occurring in a clause; and so, the metrics in Table \ref{tab:comparison} are likely somewhat conservative. This pattern may further suggest that, while having extra context for annotating complex semantic phenomena is always preferable, we still capture useful information by annotating only isolated sentences.

\vspace{-1mm}

\paragraph{Qualitative comparison} Figure \ref{fig:comparison} shows the mean normalized value for each property in our framework broken out by clause type. As expected, we see that episodics tend to have particular-referring arguments and predicates, while generics tend to have kind-referring arguments and non-particular predicates. Also as expected, episodics and habituals tend to refer to situations that are more dynamic than statives and generics. But while it makes sense that generics would be, on average, near zero for dynamicity---since generics can be about both dynamic and non-dynamic situations---it is less clear why statives are not more negative. This pattern may arise in some way from the fact that there is relatively lower agreement on dynamicity, as noted in \S\ref{sec:frameworkvalidation}.

%% file: tables/comparison.tex
\begin{tabular}{lccc|cc}
\toprule
\textbf{Clause type} &  \textbf{P} &  \textbf{R} &   \textbf{F} &  $\kappa_\text{mod}$ &  $\kappa_\text{ann}$ \\
\midrule
\textsc{eventive}                 &       0.68 &    0.55 &  0.61 &      0.49 &      0.74 \\
\textsc{stative}                 &       0.61 &    0.59 &  0.60 &      0.47 &      0.67 \\
\textsc{habitual} &       0.49 &    0.52 &  0.50 &      0.33 &      0.43 \\
\textsc{generic}      &       0.66 &    0.77 &  0.71 &      0.61 &      0.68 \\
\bottomrule
\end{tabular}

%% file: sections/06-bulkannotation.tex
We use our annotation framework to collect annotations of predicates and arguments on UD-EWT using the PredPatt system---thus yielding the Universal Decompositional Semantics--Genericity (UDS-G) dataset. Using UD-EWT in conjunction with PredPatt has two main advantages over other similar corpora: (i) UD-EWT contains text from multiple genres---not just newswire---with gold standard Universal Dependency parses; and (ii) there are now a wide variety of other semantic annotations on top of UD-EWT that use the PredPatt standard \citep{white_universal_2016,rudinger_neural_2018, vashishtha2019fine}.

\paragraph{Predicate and argument extraction} PredPatt identifies 34,025 predicates and 56,246 arguments of those predicates from 16,622 sentences. Based on analysis of the data from our validation study (\S\ref{sec:frameworkvalidation}) and other pilot experiments (not reported here), we developed a set of heuristics for filtering certain tokens that PredPatt identifies as predicates and arguments, either because we found that there was little variability in the label assigned to particular subsets of tokens---e.g.\ pronominal arguments, such as \textit{I}, \textit{we}, \textit{he}, \textit{she}, etc., are almost always labeled particular, non-kind, and non-abstract (with the exception of \textit{you} and \textit{they}, which can be kind-referring)---or because it is not generally possible to answer questions about those tokens---e.g.\ adverbial predicates are excluded. Based on these filtering heuristics, we retain 37,146 arguments and 33,114 predicates for annotation. Table \ref{tab:all_corp} compares these numbers against the resources described in \S\ref{sec:background}.

\begin{table}[t]
\small
\centering
\input{tables/corpuses.tex}
\caption{Survey of genericity annotated corpora for English, including our new corpus (in bold).}
\label{tab:all_corp}
\end{table}

\paragraph{Annotators} 482 annotators were recruited from Amazon Mechanical Turk to annotate arguments; and 438 annotators were recruited to annotate predicates. Arguments and predicates in the UD-EWT validation and test sets were annotated by three annotators each; and those in the UD-EWT train set were annotated by one each. All annotations were performed in batches of 10.

\paragraph{Annotation normalization} We use the annotation normalization procedure described in \S\ref{sec:comparison}, fit separately to our train and development splits, on the one hand, and our test split, on the other.

%% file: tables/corpuses.tex
\begin{tabular}{lllr}
	\toprule
    \textbf{Corpus} & \textbf{Level} & \textbf{Scheme} & \textbf{Size} \\
    \midrule
    ACE-2 & \multirow{2}{*}{NP} & \multirow{2}{*}{multi-class} & \multirow{2}{*}{40,106} \\
    ACE-2005 &  &  &  \\
    \midrule
    \multirow{2}{*}{ECB+} & Arg. & multi-class & 12,540 \\
    & Pred. & multi-class & 14,884 \\
    \midrule
    CFD & NP & multi-class & 3,422 \\
    \midrule
    Matthew et al & clause & multi-class & 1,052 \\
    \midrule
	ARRAU & NP & multi-class & 91,933 \\
    \midrule
    \multirow{2}{*}{SitEnt} & Topic & multi-class & \multirow{2}{*}{40,940} \\
     & Clause & multi-class & \\
    \midrule
    \multirow{2}{*}{RED}  & Arg. & multi-class & 10,319 \\
     & Pred. & multi-class & 8,731 \\     
    \midrule
    \multirow{2}{*}{\textbf{UDS-G}}  & \textbf{Arg.} & \textbf{multi-label} & \textbf{37,146} \\
     & \textbf{Pred.} & \textbf{multi-label} & \textbf{33,114} \\
    \bottomrule
\end{tabular}

%% file: sections/07-exploratoryanalysis.tex
\begin{figure*}[t]
\centering
\includegraphics[width=\columnwidth]{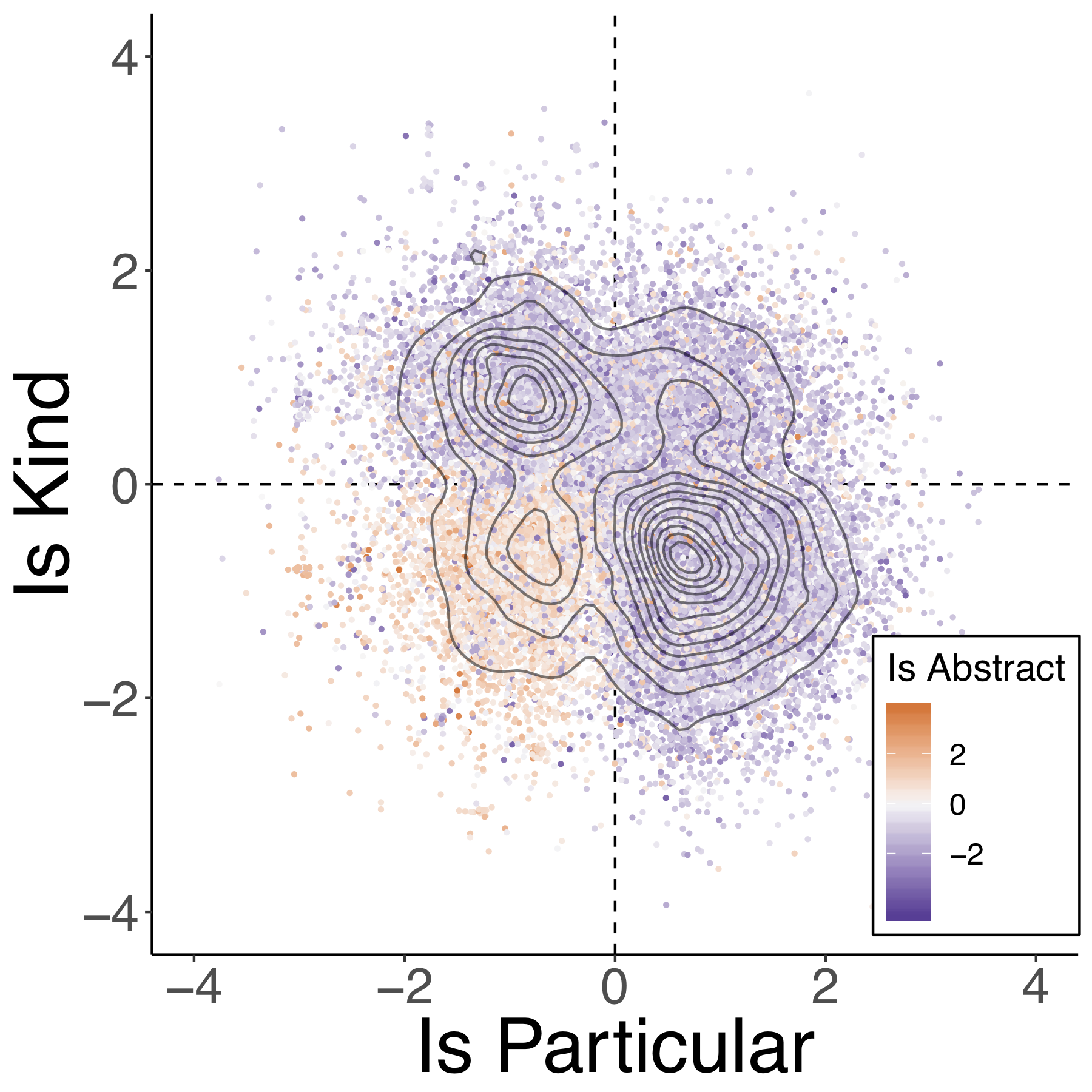}
\includegraphics[width=\columnwidth]{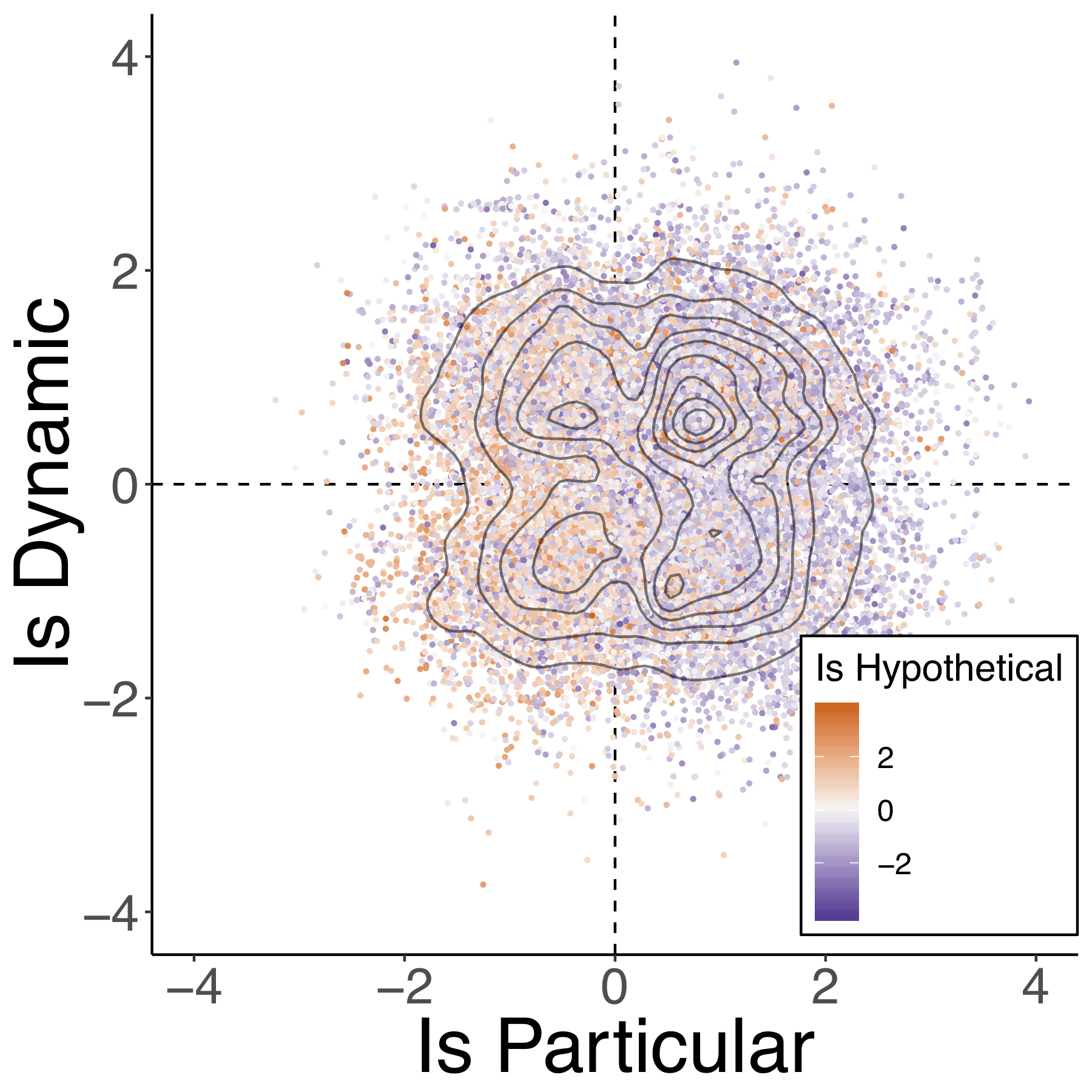}
\vspace{-4mm}
\caption{Distribution of normalized annotations in argument (left) and predicate (right) protocols.}
\label{fig:normalized}
\end{figure*}

Before presenting models for predicting our properties, we conduct an exploratory analysis to demonstrate that the properties of the dataset relate to other token- and type-level semantic properties in intuitive ways. Figure \ref{fig:normalized} plots the normalized ratings for the argument (left) and predicate (right) protocols. Each point corresponds to a token and the density plots visualize the number of points in a region. 

\paragraph{Arguments} We see that arguments have a slight tendency (Pearson correlation $\rho\!=\!{-}0.33$) to refer to either a kind or a particular---e.g.\ \textit{place} in \ref{ex:only-part} falls in the lower right quadrant (particular-referring) and \textit{transportation} in \ref{ex:only-kind} falls in the upper left quadrant (kind-referring)---though there are a not insignificant number of arguments that refer to something that is both---e.g.\ \textit{registration} in \ref{ex:part-kind} falls in the upper right quadrant.

\ex. I think this \underline{place} is probably really great especially judging by the reviews on here . \label{ex:only-part}

\ex. What made it perfect was that they offered \underline{transportation} so that[...] \label{ex:only-kind}

\ex. Some places do the \underline{registration} right at the hospital[...] \label{ex:part-kind}

We also see that there is a slight tendency for arguments that are neither particular-referring ($\rho\!=\!-0.28$) nor kind-referring ($\rho\!=\!-0.11$) to be abstract-referring---e.g.\ \textit{power} in \ref{ex:only-abs} falls in the lower left quadrant (only abstract-referring)---but that there are some arguments that refer to abstract particulars and some that refer to abstract kinds---e.g.\ both \textit{reputation} \ref{ex:part-abs} and \textit{argument} \ref{ex:kind-abs} are abstract, but \textit{reputation} falls in the lower right quadrant, while \textit{argument} falls in the upper left.

\ex.  \underline{Power} be where power lies. \label{ex:only-abs}

\ex. Meanwhile, his \underline{reputation} seems to be improving, although Bangs noted a ``pretty interesting social dynamic.'' \label{ex:part-abs}

\ex. The Pew researchers tried to transcend the economic \underline{argument}. \label{ex:kind-abs}

\paragraph{Predicates} We see that there is effectively no tendency ($\rho\!=\!0.00$) for predicates that refer to particular situations to refer to dynamic events---e.g.\ \textit{faxed} in the \ref{ex:part-dyn} falls in the upper right quadrant (particular- and dynamic-referring), while \textit{available} in \ref{ex:part-notdyn} falls in the lower right quadrant (particular- and non-dynamic-referring).

\ex. I have \underline{faxed} to you the form of Bond[...]\label{ex:part-dyn}

\ex. is gare montparnasse storage still \underline{available}? \label{ex:part-notdyn}

But we do see that there is a slight tendency ($\rho\!=\!{-}0.25$) for predicates that are hypothetical-referring not to be particular-referring---e.g.\ \textit{knows} in \ref{ex:hyp-part-a} and \textit{do} in \ref{ex:hyp-part-b} are hypotheticals in the lower left.

\ex. 
\a. Who \underline{knows} what the future might hold , and it might be expensive ? \label{ex:hyp-part-a}
\b. I have tryed to give him water but he wont take it..what should i \underline{do}? \label{ex:hyp-part-b}

%% file: sections/08-models.tex
We consider two forms of predicate and argument representations to predict the three attributes in our framework: hand-engineered features and learned features. For both, we contrast both type-level information and token-level information.

\vspace{-1mm}

\paragraph{Hand-engineered features} We consider five sets of type-level hand-engineered features.

\begin{enumerate}[itemsep=0pt]
\item \textit{Concreteness} Concreteness ratings for root argument lemmas in the argument protocol from the concreteness database \citep{brysbaert_concreteness_2014} and the mean, maximum, and minimum concreteness rating of a predicate's arguments in the predicate protocol.

\item \textit{Eventivity} Eventivity and stativity ratings for the root predicate lemma in the predicate protocol and the predicate head of the root argument in the argument protocol from the LCS database \citep{dorr1993machine}.

\item \textit{VerbNet} Verb classes from VerbNet \citep{schuler_verbnet:_2005} for root predicate lemmas.

\item \textit{FrameNet} Frames evoked by root predicate lemmas in the predicate protocol and for both the root argument lemma and its predicate head in the argument protocol from FrameNet \citep{baker_berkeley_1998}.

\item \textit{WordNet} The union of WordNet \citep{fellbaum_wordnet:_1998} \textit{supersenses} \citep{ciaramita_supersense_2003} for all WordNet senses the root argument or predicate lemmas can have.
\end{enumerate}

\noindent And we consider two sets of token-level hand-engineered features.

\begin{enumerate}[itemsep=0pt]

\item \textit{Syntactic features} POS tags, UD morphological features, and governing dependencies were extracted using PredPatt for the predicate/argument root and all of its dependents.

\item \textit{Lexical features} Function words---determiners, modals, auxiliaries---in the dependents of the arguments and predicates.
\end{enumerate}

\vspace{-1mm}

\paragraph{Learned features} For our type-level learned features, we use the 42B uncased GloVe embeddings for the root of the annotated predicate or argument \citep{pennington_glove:_2014}. For our token-level learned features, we use 1,024-dimensional ELMo embeddings \citep{peters_deep_2018}.\ To obtain the latter, the UD-EWT sentences are passed as input to the ELMo three-layered biLM, and we extract the output of all three hidden layers for the root of the annotated predicates and arguments, giving us 3,072-dimensional vectors for each.

\vspace{-1mm}

\paragraph{Labeling models} For each protocol, we predict the three normalized properties corresponding to the annotated token(s) using different subsets of the above features.\ The feature representation is used as the input to a multilayer perceptron (MLP) with ReLU nonlinearity and L1 loss.\ The number of hidden layers and their sizes are hyperparameters that we tune on the development set.

\vspace{-1mm}

\paragraph{Implementation}

For all experiments, we use stochastic gradient descent to train the MLP parameters with the Adam optimizer \cite{kingma2014adam}, using the default learning rate in pytorch ($10^{-3}$).\ We performed ablation experiments on the 4 major classes of features discussed above.

\paragraph{Hyperparameters} For each of the ablation experiments, we ran a hyperparameter grid search over hidden layer sizes (one or two hidden layers with sizes $h_1, h_2 \in$ $\left\{512, 256, 128, 64, 32\right\}$; $h_2$ at most half of $h_1$), L2 regularization penalty $l \in$ $\left\{0, 10^{-5}, 10^{-4}, 10^{-3}\right\}$, and the dropout probability $d \in$ $\left\{0.1, 0.2, 0.3, 0.4, 0.5\right\}$.

\paragraph{Development} For all models, we train for at most 20 epochs with early stopping. At the end of each epoch, the L1 loss is calculated on the development set, and if it is higher than the previous epoch, we stop training, saving the parameter values from the previous epoch.

\paragraph{Evaluation} Consonant with work in event factuality prediction, we report Pearson correlation ($\rho$) and proportion of mean absolute error (MAE) explained by the model, which we refer to as R1 on analogy with the variance explained R2 = $\rho^2$.
\vspace{-2mm}
\begin{align*}
\text{R1} &= 1-\frac{\text{MAE}^p_\text{model}}{\text{MAE}^p_\text{baseline}}    
\end{align*}

\vspace{-1mm}

\noindent where $\text{MAE}^p_\text{baseline}$ is always guessing the median for property $p$. We calculate R1 across properties (wR1) by taking the mean R1 weighted by the MAE for each property.

These metrics together are useful, since $\rho$ tells us how similar the predictions are to the true values, ignoring scale, and R1 tells us how close the predictions are to the true values, after accounting for variability in the data. We focus mainly on differences in relative performance among our models, but for comparison, state-of-the-art event factuality prediction systems obtain $\rho\!\approx0.77$ and R1 $\approx0.57$ for predicting event factuality on the predicates we annotate \citep{rudinger_neural_2018}. 

%% file: sections/09-results.tex
\begin{table*}[t]
\small
\centering
\input{tables/all_ablations_test.tex}
\caption{Correlation ($\rho$) and MAE explained (R1) on test split for argument (top) and predicate (bottom) protocols. Bolded numbers give the best result in the column; the models highlighted in blue are the ones analyzed in \S\ref{sec:analysis}.}
\label{tab:results}
\end{table*}

Table \ref{tab:results} contains the results on the test set for both the argument (top) and predicate (bottom) protocols. We see that (i) our models are generally better able to predict referential properties of arguments than those of predicates; (ii) for both predicates and arguments, contextual learned representations contain most of the relevant information for both arguments and predicates, though the addition of hand-engineered features can give a slight performance boost, particularly for the predicate properties; and (iii) the proportion of absolute error explained is significantly lower than what we might expect from the variance explained implied by the correlations. We discuss (i) and (ii) here, deferring discussion of (iii) to \S\ref{sec:analysis}.

\vspace{-1mm}

\paragraph{Argument properties} 

While type-level hand-engineered and learned features perform relatively poorly for properties such as \textsc{is.particular} and \textsc{is.kind} for arguments, they are able to predict \textsc{is.abstract} relatively well compared to the models with all features. The converse of this also holds: token-level hand-engineered features are better able to predict \textsc{is.particular} and \textsc{is.kind}, but perform relatively poorly on their own for \textsc{is.abstract}. 

This seems likely to be a product of abstract reference being fairly strongly associated with particular lexical items, while most arguments can refer to particulars and kinds (and which they refer to is context-dependent). And in light of the relatively good performance of contextual learned features alone, it suggests that these contextual learned features---in contrast to the hand-engineered token-level features---are able to use this information coming from the lexical item. 

Interestingly, however, the models with both contextual learned features (ELMo) and hand-engineered token-level features perform slightly better than those without the hand-engineered features across the board, suggesting that there is some (small) amount of contextual information relevant to generalization that the contextual learned features are missing. This performance boost may be diminished by improved contextual encoders, such as BERT \citep{devlin-etal-2019-bert}.  

\vspace{-1mm}

\paragraph{Predicate properties} 

We see a pattern similar to the one observed for the argument properties mirrored in the predicate properties: while type-level hand-engineered and learned features perform relatively poorly for properties such as \textsc{is.particular} and \textsc{is.hypothetical}, they are able to predict \textsc{is.dynamic} relatively well compared to the models with all features. The converse of this also holds: token-level hand-engineered features are better able to predict \textsc{is.particular} and \textsc{is.hypothetical}, but perform relatively poorly on their own for \textsc{is.dynamic}.

One caveat here is that, unlike for \textsc{is.abstract}, type-level learned features (GloVe) alone perform quite poorly for \textsc{is.dynamic}, and the difference between the models with only type-level hand-engineered features and the ones with only token-level hand-engineered features is less stark for \textsc{is.dynamic} than for \textsc{is.abstract}. This may suggest that, though \textsc{is.dynamic} is relatively constrained by the lexical item, it may be more contextually determined than \textsc{is.abstract}. Another major difference between the argument properties and the predicate properties is that \textsc{is.particular} is much more difficult to predict than \textsc{is.hypothetical}.\ This contrasts with \textsc{is.particular} for arguments, which is easier to predict than \textsc{is.kind}.

%% file: tables/all_ablations_test.tex
\begin{tabular}{cccccccccccc}
\toprule
& \multicolumn{4}{c}{Feature sets} & \multicolumn{2}{c}{\textbf{Is.Particular}} &
\multicolumn{2}{c}{\textbf{Is.Kind}} &
\multicolumn{2}{c}{\textbf{Is.Abstract}} & \textbf{All}\\
\cmidrule(lr){2-5}
\cmidrule(lr){6-7}
\cmidrule(lr){8-9}
\cmidrule(lr){10-11}
& \small{Type} & \small{Token} &
\small{GloVe} &
\small{ELMO} &
$\rho$ & R1 & $\rho$ & R1 & $\rho$ & R1 & wR1\\
\midrule
{\multirow{9}{*}{\rotatebox{90}{ARGUMENT}}}
& + & - & - & - & 42.4 & 7.4 & 30.2 & 4.9 & 51.4 & 11.7 & 8.1 \\
& - & + & - & - & 50.6 & 13.0 & 41.5 & 8.8 & 33.8 & 4.8 & 8.7 \\
& - & - & + & - & 44.5 & 8.3 & 33.4 & 4.6 & 45.2 & 7.7 & 6.9 \\
& - & - & - & + & 57.5 & 17.0 & 48.1 & 13.3 & 55.7 & 14.9 & 15.1 \\
& + & + & - & - & 55.3 & 14.1 & 46.2 & 11.6 & 52.6 & 13.0 & 12.9 \\
& - & + & - & + & \textbf{58.6} & 15.6 & 48.6 & \textbf{13.7} & \textbf{56.8} & 14.2 & 14.5 \\
& + & + & - & + & 58.3 & 16.3 & 47.8 & 13.2 & 56.3 & \textbf{15.2} & 14.9 \\
\rowcolor{blue!10}& + & + & + & + & 58.1 & \textbf{17.0} & \textbf{48.9} & 13.2 & 56.1 & 15.1 & \textbf{15.1} \\
\midrule
&&&&& \multicolumn{2}{c}{\textbf{Is.Particular}} & \multicolumn{2}{c}{\textbf{Is.Hypothetical}} & \multicolumn{2}{c}{\textbf{Is.Dynamic}} & \\
\cmidrule(lr){6-7}
\cmidrule(lr){8-9}
\cmidrule(lr){10-11}
\multirow{9}{*}{\rotatebox{90}{PREDICATE}}
& + & - & - & - & 14.0 & 0.8 & 13.4 & 0.0 & 32.5 & 5.6 & 2.0 \\
& - & + & - & - & 22.3 & 2.8 & 37.7 & 7.3 & 31.7 & 5.1 & 5.1 \\
& - & - & + & - & 20.6 & 2.2 & 23.4 & 2.4 & 29.7 & 4.6 & 3.0 \\
& - & - & - & + & 26.2 & 3.6 & 43.1 & 10.0 & 37.0 & 6.8 & 6.8 \\
& - & - & + & + & 26.8 & 4.0 & 42.8 & 8.9 & 37.3 & 7.3 & 6.7 \\
& + & + & - & - & 24.0 & 3.3 & 37.9 & 7.6 & 37.1 & 7.6 & 6.1 \\
& - & + & - & + & \textbf{27.4} & 4.1 & 43.3 & 10.1 & \textbf{38.6} & \textbf{7.8} & \textbf{7.4} \\
& + & - & - & + & 27.1 & 4.0 & 43.0 & 10.1 & 37.5 & 7.6 & 7.2 \\
\rowcolor{blue!10} & + & + & + & + & 26.8 & \textbf{4.1} & \textbf{43.5} & \textbf{10.3} & 37.1 & 7.2 & 7.2 \\
\bottomrule
\end{tabular}

%% file: sections/10-analysis.tex
\begin{figure*}[t]
    \centering
    \includegraphics[width=2\columnwidth]{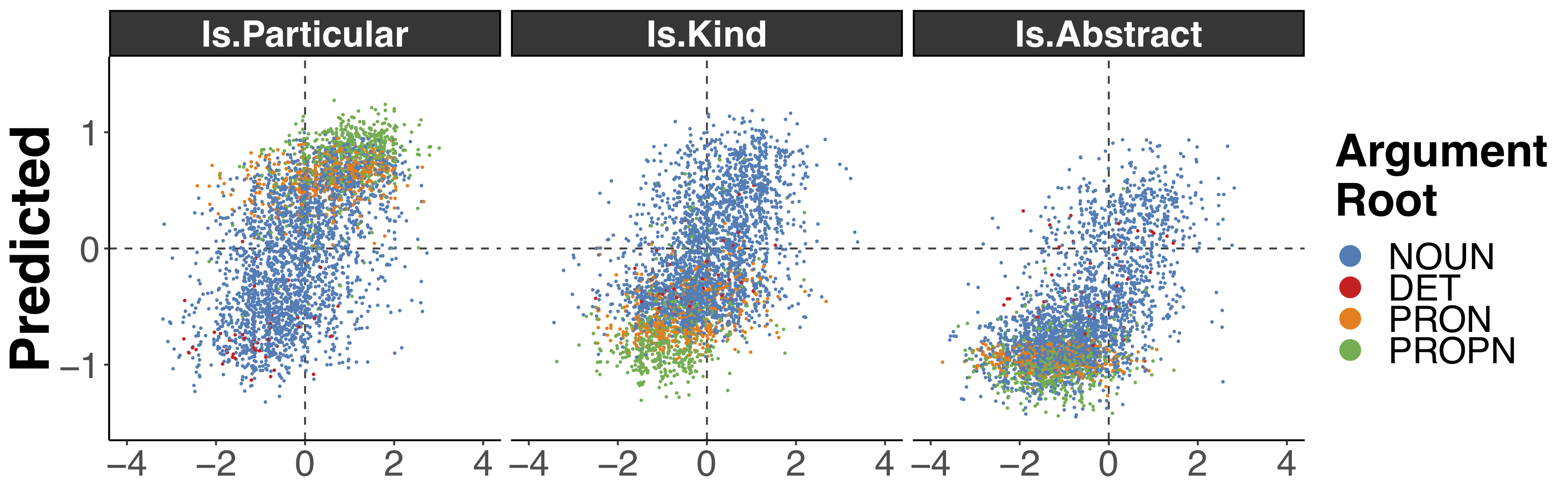}
    \includegraphics[width=2\columnwidth]{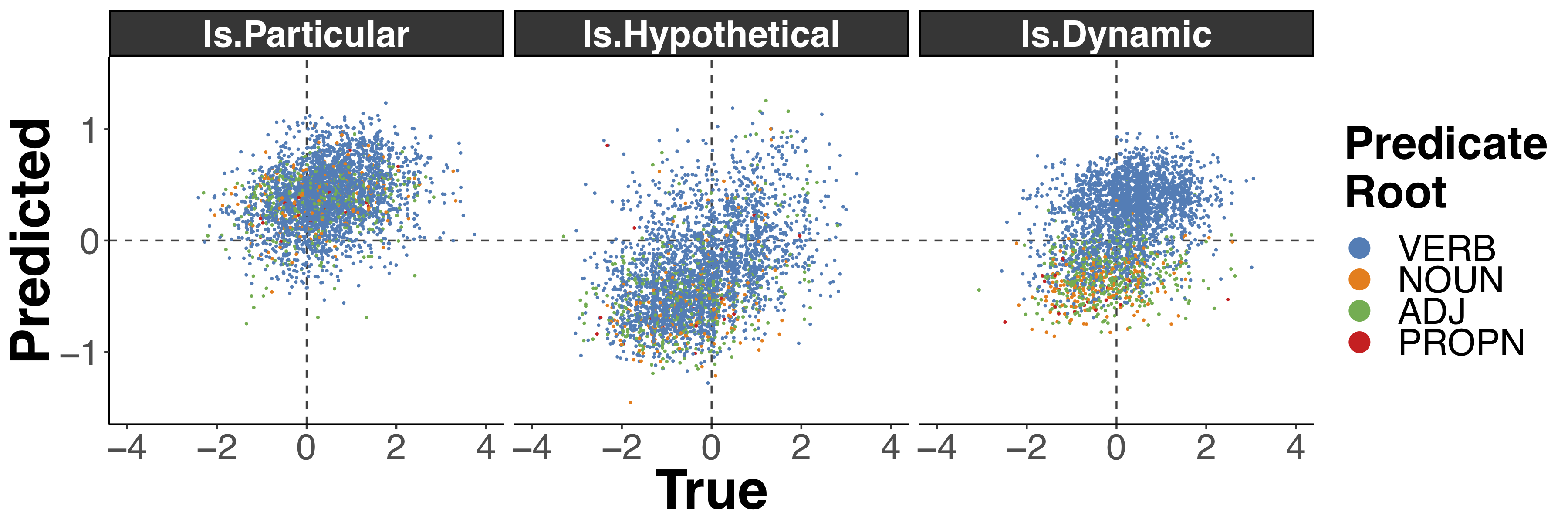}
    \vspace{-4mm}
    \caption{True (normalized) property values for argument (top) and predicate (bottom) protocols in the development set plotted against values predicted by models highlighted in blue in Table \ref{tab:results}.}
    \label{fig:true_v_predicted}
\end{figure*}

Figure \ref{fig:true_v_predicted} plots the true (normalized) property values for the argument (top) and predicate (bottom) protocols from the development set against the values predicted by the models highlighted in blue in Table \ref{tab:results}. Points are colored by the part-of-speech of the argument or predicate root.

We see two overarching patterns. First, our models are generally reluctant to predict values outside the $[-1, 1]$ range, despite the fact that there are not an insignificant number of true values outside this range. This behavior likely contributes to the difference we saw between the $\rho$ and R1 metrics, wherein R1 was generally worse than we would expect from $\rho$. This pattern is starkest for \textsc{is.particular} in the predicate protocol, where predictions are nearly all constrained to $[0,1]$.

Second, the model appears to be heavily reliant on part-of-speech information---or some semantic information related to part-of-speech---for making predictions. This behavior can be seen in the fact that, though common noun-rooted arguments get relatively variable predictions, pronoun- and proper noun-rooted arguments are almost always predicted to be particular, non-kind, non-abstract; and though verb-rooted predicates also get relatively variable predictions, common noun-, adjective-, and proper noun-rooted predicates are almost always predicted to be non-dynamic.

\paragraph{Argument protocol}

Proper nouns tend to refer to particular, non-kind, non-abstract entities, but they can be kind-referring, which our models miss: \textit{iPhone} in \ref{ex:part-abs-b} and \textit{Marines} in \ref{ex:part-abs-a} were predicted to have low kind score and high particular score, while annotators label these arguments as non-particular and kind-referring.

\ex. \underline{The US Marines} took most of Fallujah Wednesday, but still face[...] \label{ex:part-abs-a}

\ex. I'm writing an essay...and I need to know if \underline{the iPhone} was the first Smart Phone. \label{ex:part-abs-b}

This similarly holds for pronouns. As mentioned in \S\ref{sec:datacollection}, we filtered out several pronominal arguments, but certain pronouns---like \textit{you}, \textit{they}, \textit{yourself},\textit{ themselves}---were not filtered because they can have both particular- and kind-referring uses. Our models fail to capture instances where pronouns are labeled kind-referring---e.g.\ \textit{you} in \ref{ex:pron-a} and \ref{ex:pron-b}---consistently predicting low \textsc{is.kind} scores, likely because they are rare in our data.

\ex. I like Hayes Street Grill....another plus, it's right by Civic Center, so \underline{you} can take a romantic walk around the Opera House, City Hall, Symphony Auditorium[...] \label{ex:pron-a}

\ex. What would happen if \underline{you} flew the flag of South Vietnam in Modern day Vietnam? \label{ex:pron-b}

\noindent This behavior is not seen with common nouns: the model correctly predicts common nouns in certain contexts as non-particular, non-abstract, and kind-referring---e.g.\ \textit{food} in \ref{ex:part-abs-c} and \textit{men} in \ref{ex:part-abs-d}. 

\ex. Kitchen puts out good \underline{food}[...]\label{ex:part-abs-c}

\ex. just saying most \underline{men} suck! \label{ex:part-abs-d}

\paragraph{Predicate protocol}

As in the argument protocol, general trends associated with part-of-speech are exaggerated by the model. We noted in \S\ref{sec:preliminaryanalysis} that annotators tend to annotate hypothetical predicates as non-particular and vice-versa ($\rho\!=\!{-}0.25$), but the model's predictions are anti-correlated to a much greater extent ($\rho\!=\!{-}0.79$). For example, annotators are more willing to say a predicate can refer to particular, hypothetical situations \ref{ex:part-hyp-b} or a non-particular, non-hypothetical situation \ref{not-part-not-hyp-b}.

\ex. \underline{Read} the entire article[...]\label{ex:part-hyp-b}

\ex. it \underline{s illegal} to sell stolen property, even if you don't know its stolen. \label{not-part-not-hyp-b}

The model also had a bias towards particular predicates referring to dynamic predicates($\rho\!=\!0.34$)---a correlation not present among annotators. For instance, \textit{is closed} in \ref{ex:part-dyn-a} was annotated as particular but non-dynamic but predicted by the model to be particular and dynamic; and \textit{helped} in \ref{ex:part-dyn-b} was annotated as non-particular and dynamic, but the model predicted particular and dynamic.

\ex. library \underline{is closed} \label{ex:part-dyn-a}

\ex. I have a new born daughter and she \underline{helped} me with a lot. \label{ex:part-dyn-b}

%% file: sections/11-conclusion.tex
We have proposed a novel semantic framework for modeling linguistic expressions of generalization as combinations of simple, real-valued referential properties of predicates and their arguments. We used this framework to construct a dataset covering the entirety of the Universal Dependencies English Web Treebank and probed the ability of both hand-engineered and learned type- and token-level features to predict the annotations in this dataset.

%% file: sections/acknowledgments.tex
We would like to thank three anonymous reviewers and Chris Potts for useful comments on this paper as well as Scott Grimm and the FACTS.lab at the University of Rochester for useful comments on the framework and protocol design. This research was supported by the University of Rochester, JHU HLTCOE, and DARPA AIDA.  The U.S. Government is authorized to reproduce and distribute reprints for Governmental purposes. The views and conclusions contained in this publication are those of the authors and should not be interpreted as representing official policies or endorsements of DARPA or the U.S. Government.